\documentclass[letterpaper]{article} 
\usepackage[preprint]{aaai2027}  
\usepackage[hyphens]{url}  
\usepackage{graphicx} 
\usepackage{natbib}  
\usepackage{caption} 
\usepackage{algorithm}
\usepackage{algorithmic}

\usepackage{makecell}
\usepackage{amssymb}
\usepackage{amsmath}
\usepackage{multirow}
\usepackage[table]{xcolor}
\definecolor{sectiongray}{gray}{0.92}
\usepackage{booktabs}
\newcolumntype{V}{%
  !{\hspace{0.5em}\vrule width 0.45pt\hspace{0.5em}}%
}

\usepackage{newfloat}
\usepackage{listings}
\DeclareCaptionStyle{ruled}{labelfont=normalfont,labelsep=colon,strut=off} 
\floatstyle{ruled}
\newfloat{listing}{tb}{lst}{}
\floatname{listing}{Listing}

\usepackage{booktabs}

\title{Beyond Relative Geometry: Metric-Aware Geometry Perception for Robotics}

\author{
    Fengjun Zhong\textsuperscript{\rm 1}, Congjia Chen\textsuperscript{\rm 1}, Zhaoxu Liu\textsuperscript{\rm 1}, Jinyang Du\textsuperscript{\rm 1}, Yuchen Gong\textsuperscript{\rm 1}, Enqi Mao\textsuperscript{\rm 1}, Ruihao Gong\textsuperscript{\rm 1}, ShuJie Wang\textsuperscript{\rm 2}, Xianglong Liu\textsuperscript{\rm 1}\corresponding, Zhongliang Qiao\textsuperscript{\rm 2}
}

\affiliations{
    \textsuperscript{\rm 1}Beihang University, \textsuperscript{\rm 2}XiaoyuBot\\

}

\begin{document}

\maketitle

\begin{abstract}

Recent embodied models increasingly leverage geometric representations to improve spatial reasoning and robotic manipulation. However, existing reconstruction methods only reconstruct relative geometry with arbitrary scales, causing predicted object dimensions and spatial distances to vary across scenes, viewpoints, and input configurations. This inconsistency prevents geometric perception from being directly aligned with robotic actions defined on the real-world scale. To address this limitation, we propose Metric-Aware Geometry Perception (MAGP), an end-to-end, plug-and-play framework for metric geometry reconstruction that can be seamlessly integrated into robotic policies. At its core, Metric Scale Equivariant Augmentation encourages the model to reconstruct metric geometry from camera parameters and depth observations, ensuring that the reconstructed geometry follows the metric scale specified by observations. Flexible Metric Conditioning further enables MAGP to support arbitrary view counts and combinations of camera and depth inputs, improving robustness to heterogeneous robotic sensing configurations. Together, these designs produce geometrically consistent reconstructions with stable object dimensions and spatial distances across scenes and sensing conditions. Experiments on ETH3D, MegaDepth, and ScanNet++ demonstrate that MAGP maintains strong relative geometry accuracy while reducing the absolute error by over an order of magnitude, from $2.01\mathrm{m}$ to $0.07\mathrm{m}$. When integrated into multiple robotic policies, MAGP consistently improves performance on LIBERO, RoboTwin, and zero-shot LIBERO-Plus, with gains of up to 6.26\% on RoboTwin. These results demonstrate the effectiveness and generalizability of metric geometry for robotic manipulation.

\end{abstract}



\section{Introduction}




Current embodied models mainly rely on 2D visual observations, which provide rich semantics but lack explicit geometry for spatial reasoning and precise manipulation~\cite{black2025pi_, kim2024openvla}. With the rapid development of feed-forward 3D reconstruction, an increasing number of embodied models have incorporated depth, point clouds, and multi-view geometry to provide explicit spatial structures for scene understanding and action prediction~\cite{yang2026abot, yuan2025depthvla, wu2026pragmatic, ni2025vo, qu2025spatialvla, shridhar2023perceiver, zhen20243d, zhang2026gear, liu2026lift3d, ge2025vggt}. Under perspective projection, scenes with different physical dimensions and camera distances can produce indistinguishable visual observations. Existing reconstruction models therefore only recover relative geometry~\cite{wang2025moge, wang2026moge, wang2026vggt, wang2025vggt}. Such relative geometry preserves scene shape and spatial layout but lacks a consistent metric reference. As the viewpoint or camera input order changes, the same object will be reconstructed at different scales, resulting in inconsistent estimates of object dimension and spatial distance.

\begin{figure}[t]
\centering
\includegraphics[width=0.99\columnwidth]{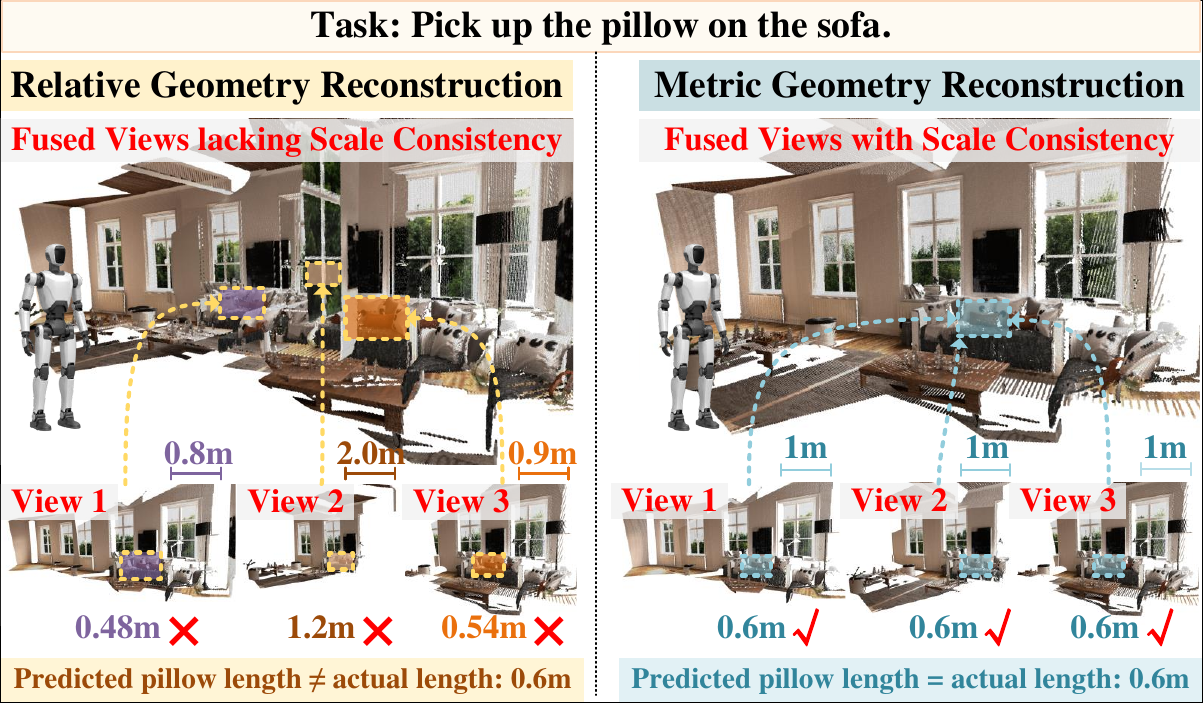}
\caption{Illustrative example of relative/metric geometry reconstruction across different views. Metric geometry preserves consistent object dimensions and spatial distances across views, whereas relative geometry exhibits view-dependent scale variations.}
\label{fig_motivation}
\end{figure}

As illustrated in Fig.~\ref{fig_motivation}, relative geometry reconstruction assigns inconsistent object dimensions to the same object across views and produces spatially misaligned multi-view geometry in a shared coordinate system. Consequently, the same robot motion has different geometric displacements under different perspectives, disrupting the correspondence between robot perception and action. This causes inaccurate target localization, miscalibrated end-effector motion, and degraded fine-grained manipulation. In contrast, metric geometry anchors all observations to a consistent metric scale, preserving consistent object dimensions and spatial relationships across views. Such a scale-aligned representation is essential for embodied intelligence, as it provides a reliable geometric basis for target localization, action calibration, motion planning, and precise manipulation.

However, achieving metric geometry reconstruction is extremely challenging. It requires reconstruction models to infer a globally consistent metric scale from metric observations, such as camera translation and depth observations. In practice, reconstruction models often rely on visual priors, including object categories, scene layouts, and appearance statistics, rather than learning the intrinsic correspondence between metric observations and scene scale. This challenge becomes more severe under flexible observation configurations, where the availability, quantity, and depth map density of metric observations vary across different robotic platforms. The model must exploit all available metric evidence effectively while remaining robust to changes in sensor configuration.


To address these challenges, we propose Metric-Aware Geometry Perception (MAGP), an end-to-end, plug-and-play metric reconstruction framework for robotics. MAGP jointly encodes multi-view images with optional camera parameters and depth observations into geometry tokens that explicitly preserve metric scale, enabling seamless integration with embodied policies. Its core, \emph{metric scale equivariant reconstruction}, is achieved through Metric Scale Equivariant Augmentation (MSEA), which perturbs metric observations to encourage scale inference from camera translation and depth rather than appearance-based priors. Fig.~\ref{fig_scale_equivariance} demonstrates that the \emph{reconstructed scene geometry matches the metric scale specified by the metric observation}. Flexible Metric Conditioning further supports arbitrary view counts and combinations of metric inputs, accommodating heterogeneous robotic sensing configurations. Finally, joint relative and metric supervision preserves fine-grained geometry while ensuring reconstruction in a consistent metric scale. 

Experiments on ETH3D, MegaDepth, and ScanNet++ demonstrate that MAGP improves both relative and metric geometry reconstruction. In the image-only setting, MAGP achieves depth and point scores of 74.23\% and 84.27\%, surpassing Depth Anything 3 by 17.41\% and 17.00\%, respectively. With camera parameters and depth observations, these scores further increase to 84.81\% and 87.34\%, while the corresponding absolute errors decrease to $0.11\mathrm{m}$ and $0.06\mathrm{m}$. When integrated into embodied policies, MAGP improves performance by up to 1.3\% on LIBERO and 6.26\% on RoboTwin, while achieving a further 3.9\% gain in zero-shot evaluation on LIBERO-Plus.





Our contributions are summarized as follows:
\begin{itemize}

    \item To our knowledge, MAGP is the first end-to-end and plug-and-play framework to provide metric geometry perception for robotic manipulation, bridging geometric reconstruction and action prediction under a unified scale.
    

    \item MAGP supports flexible combinations of images, camera parameters, and depth observations, recovering metric geometry whenever metric observations are available.

    \item Extensive experiments across three reconstruction benchmarks and three robotic manipulation benchmarks demonstrate that MAGP preserves strong relative geometry accuracy while reducing the absolute error from $2.01\mathrm{m}$ to $0.07\mathrm{m}$, and consistently improves different robotic policies by up to 6.26\%.
\end{itemize}


\section{Related Work}

\paragraph{3D Geometry Reconstruction.}

Recent 3D reconstruction has shifted from optimization-based pipelines to feed-forward models that directly infer scene geometry and camera parameters. DUSt3R~\cite{wang2024dust3r} and MASt3R~\cite{leroy2024grounding} formulate reconstruction through pairwise pointmap regression and geometric matching, while VGGT~\cite{wang2025vggt}, CUT3R~\cite{wang2025continuous}, and $\pi^3$~\cite{wang2025pi} extend this paradigm to unified multi-view modeling. Depth Anything 3~\cite{lin2025depth} introduces a depth-ray representation, MapAnything~\cite{keetha2026mapanything} supports flexible geometric observations, and VGGT-$\Omega$~\cite{wang2026vggt} improves performance through large-scale model and data scaling. Despite these advances, these methods remain focused on relative geometry, neglecting consistent metric scale recovery.


\paragraph{Geometry-Aware Robotic Manipulation.}

Recent studies have incorporated 3D geometry into embodied models through several representative designs~\cite{huang2026graphcot,xia20263dvla,liu2026lift3d,zhang2026gear}. 3DS-VLA~\cite{li20253ds} aligns 3D observations with pretrained visual features, while DepthVLA~\cite{yuan2025depthvla} introduces depth representations for spatial reasoning. VGGT-DP~\cite{ge2025vggt} injects pretrained VGGT features through a plug-and-play module, whereas ABot-M0~\cite{yang2026abot} adopts a dual-stream architecture to combine visual-language and VGGT-based geometric representations. Although these methods improve manipulation performance, their geometry only captures relative structure. In contrast, our MAGP recovers metric geometry and directly embeds metric information into geometric tokens for robotics.

\section{Challenge of Metric Scale 3D Perception}


As illustrated in Fig.~\ref{fig_motivation}, existing reconstruction models can recover accurate relative geometry, yet their predicted scales vary across views. Such inconsistency prevents reconstruction within a shared physical coordinate system and weakens the alignment between spatial perception and robotic actions. This section formalizes the learning objectives of relative and metric reconstruction and identifies the key challenge in recovering consistent scene scale.

\paragraph{Relative geometry lacks the metric precision required for reliable robotic manipulation.}
Given a sequence of $N$ images
$\mathcal{I}=\{I_i\}_{i=1}^{N}$,
each view is associated with depth $D_i$, camera intrinsics $K_i$, and extrinsics $[R_i|t_i]$.
$R_i$ and $t_i$ transform camera coordinates into the world coordinate system, a pixel
$\tilde{p}=(u,v,1)^{\mathsf T}$
with depth $D_i(u,v)$ is back-projected as
\begin{equation}
    P_i(u,v)=R_iD_i(u,v)K_i^{-1}\tilde{p}+t_i.
    \label{eq_back_projection}
\end{equation}

Aggregating valid points from all views yields the metric scene point cloud $\mathcal{P}$.

To resolve the inherent scale ambiguity of visual observations, existing reconstruction models $\phi_{\theta}(\cdot)$ commonly normalize the prediction and ground truth independently and optimize a relative geometry reconstruction objective:
\begin{equation}
    \mathcal{L}_{\mathrm{rel}}=d\left(\mathcal{N}(\hat{\mathcal{P}}),\mathcal{N}(\mathcal{P})\right),
    \label{eq_rel_loss}
\end{equation}
where $\hat{\mathcal{P}}=\phi_{\theta}(\mathcal{I})$ denotes the point cloud predicted by reconstruction model $\phi_{\theta}(\cdot)$, $\mathcal{N}(\cdot)$ denotes scene-level scale normalization and $d(\cdot)$ measures geometric discrepancy. Since $\mathcal{N}(s\mathcal{P})=\mathcal{N}(\mathcal{P})$
for any $s>0$, the model is only required to predict
\begin{equation}
    \hat{\mathcal{P}}=s\mathcal{P},s \sim \phi_{\theta} (\mathcal{I}),
\end{equation}
where $s$ is an unknown, sample-dependent scale factor.

Consequently, the reconstructed scale varies with view selection, camera motion, and visual content, causing identical geometric distances to correspond to different real-world measurements. This ambiguity is particularly detrimental to robotic manipulation, where geometry must calibrate end-effector displacement, gripper width, and motion trajectories. Relative geometry can indicate where to act, but cannot reliably determine how far to move in physical space.





\begin{figure}[t]
\centering
\includegraphics[width=0.99\columnwidth]{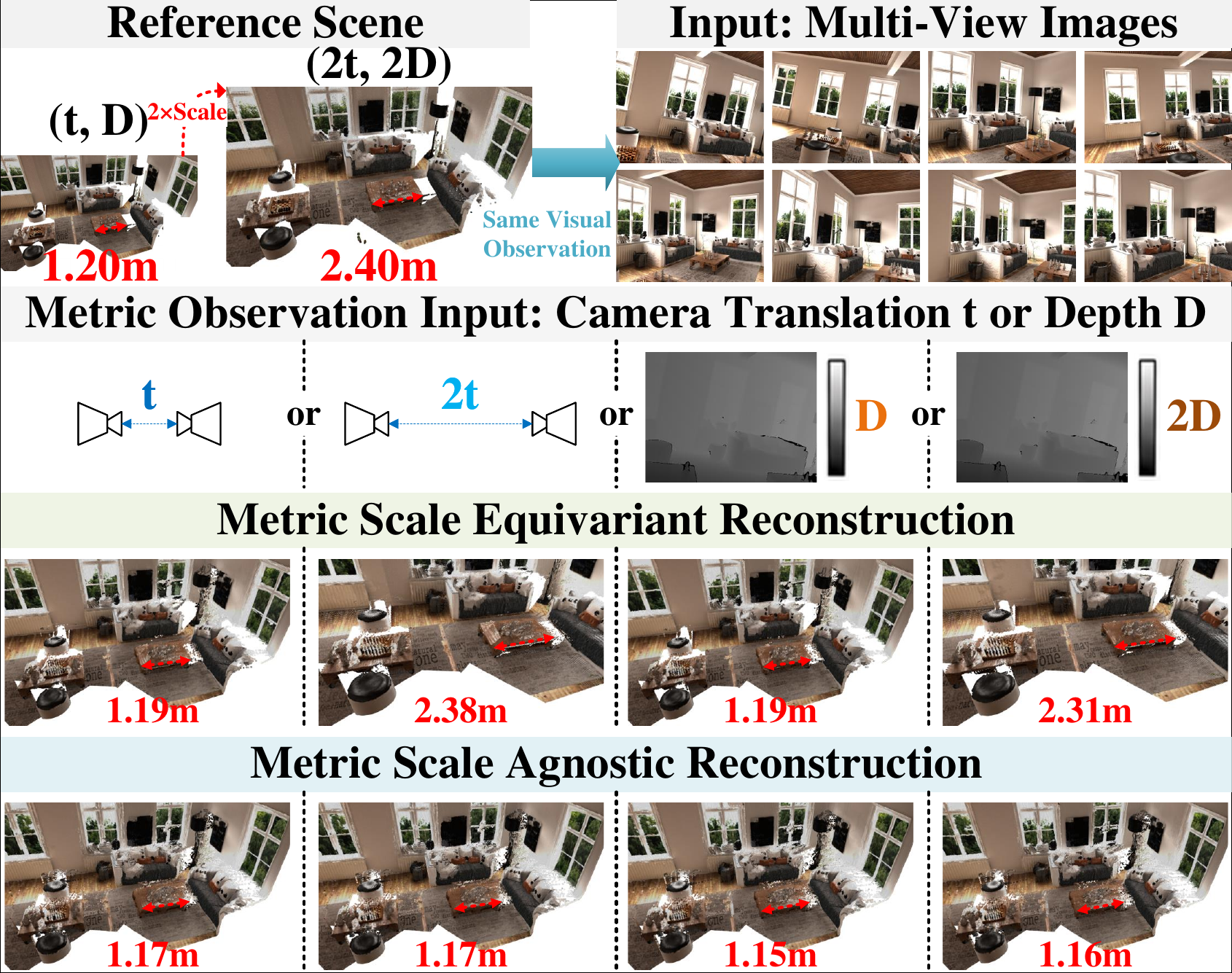}
\caption{Scenes with scaled camera translation and depth, $(\mathbf{t}, \mathcal{D})\rightarrow(2\mathbf{t}, 2\mathcal{D})$, can yield same multi-view images. Given these metric observations, metric scale equivariant reconstruction follows the corresponding scale change, whereas metric scale agnostic reconstruction remains unchanged.}
\label{fig_scale_equivariance}
\end{figure}

\paragraph{Metric scale should be grounded in camera translation and depth observations.}
Metric scale is determined by two metric observations. Camera translation $t$ provides a global metric baseline that, together with camera intrinsics, rotations, and cross-view correspondences, resolves the scale ambiguity between camera motion and scene structure through triangulation. Depth observations $\mathcal{D}$ provide local metric anchors by specifying physical distances along camera rays.

A straightforward metric reconstruction approach is to remove scene-level normalization and directly regress the scene geometry in physical units:
\begin{equation}
    \hat{\mathcal{P}}=s\mathcal{P}, s \in \mathbb{R}^+.
\end{equation}
where $\hat{\mathcal{P}}=\phi_{\theta}(\mathcal{I}, \mathcal{K}, \mathcal{R}, t, \mathcal{D})$. Accordingly, the reconstruction model is required to be optimized by

\begin{equation}
    \mathcal{L}_{\mathrm{metric}}=d\left(\hat{\mathcal{P}},\mathcal{P}\right).
    \label{eq_metric_loss}
\end{equation}



However, we find that \textbf{metric supervision $\mathcal{L}_{\mathrm{metric}}$ alone is insufficient for reliable metric geometry modeling.} As illustrated by the metric scale agnostic reconstruction in Fig.~\ref{fig_scale_equivariance} and corroborated by the results in Table~\ref{tab_compare_msea}, reconstruction models will overlook the metric information provided by camera translation and depth. Instead, they tend to infer scene scale from visual correlations, such as object appearance, semantic categories, and scene layout. Consequently, the predicted scale becomes primarily dependent on the visual observations, i.e., $s \sim \phi_{\theta}(\mathcal{I})$. This challenge is further exacerbated in practical robotic systems, where the availability of metric observations varies, making it difficult for models to extract reliable metric information from sparse and partially available observations.


Metric reconstruction should explicitly ground scene scale in the available metric observations rather than infer it from visual appearance. The reconstructed geometry should be formulated as
\begin{equation}
    \hat{\mathcal{P}}=s\mathcal{P},s \sim \phi_{\theta}(t,\mathcal{D}),
\end{equation}
where $s$ denotes the metric scale inferred from camera translation and depth observations. Accordingly, scaling the available metric observations by $\alpha>0$ should proportionally scale the reconstruction, which we call \emph{metric scale equivariant reconstruction}:
\begin{equation}
    \phi_{\theta}\left(\mathcal{I},\mathcal{K},\mathcal{R},\alpha t,\alpha\mathcal{D}\right)=\alpha\phi_{\theta}\left(\mathcal{I},\mathcal{K},\mathcal{R},t,\mathcal{D}\right).
    \label{eq_target}
\end{equation}

This attribute ensures that the reconstructed geometry follows the scale explicitly encoded by the metric observations.

\begin{figure*}[t]
\centering
\includegraphics[width=0.99\linewidth]{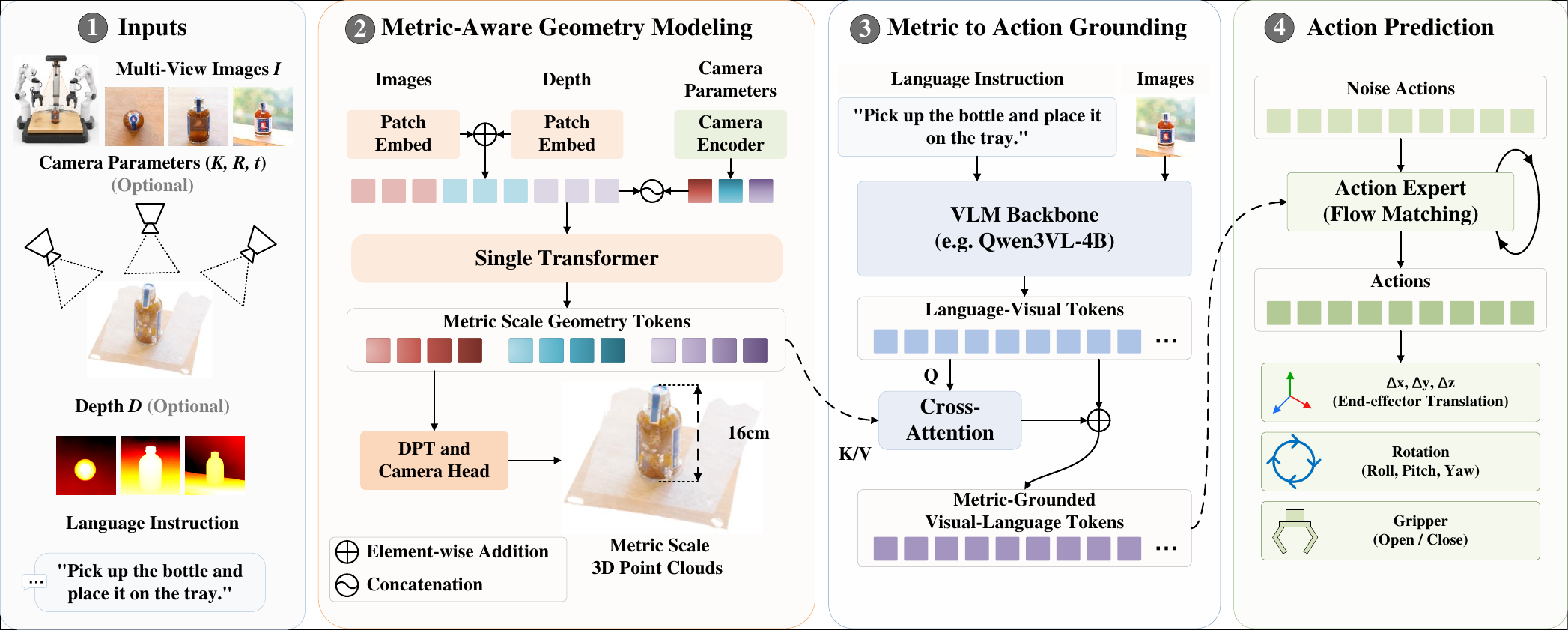}
\caption{Architecture schematic of MAGP. MAGP encodes multi-view images, optional camera parameters, and depth observations into metric geometry tokens using a single transformer. These tokens ground visual-language representations through residual cross-attention, providing metric scale calibrated geometry for action prediction.}
\label{fig_architecture}
\end{figure*}


\section{Metric-Aware 3D Perception for Robotics}


Building on metric scale equivariant reconstruction, we propose MAGP, a unified framework that produces metric geometry for robotics. MAGP combines Metric Scale Equivariant Augmentation, Flexible Metric Conditioning, and joint relative and metric scale supervision to align reconstruction scale with metric observations, support flexible observation combinations, and preserve accurate geometric structure.




\subsection{Architecture Design}

As illustrated in Fig.~\ref{fig_architecture}, MAGP is an end-to-end metric geometry reconstruction model with a single-transformer backbone. It jointly encodes multi-view images, optional camera parameters, and depth observations into metric geometry tokens that capture scale-consistent scene structure, object dimensions, and spatial distances. These tokens are then integrated into the embodied policy, providing scale-consistent geometric representations for robotics.

\paragraph{Metric-Aware Geometry Modeling.}
MAGP employs a single-transformer backbone for unified multi-view geometric modeling. Given $N$ images $\{I_i\}_{i=1}^{N}$, each view is converted into patch tokens by the patch embedding $\varepsilon_I(\cdot)$: $z_i^I = \varepsilon_I(I_i)$. The early transformer layers perform within-view self-attention to capture view-specific structure, while the subsequent layers alternate between within-view and cross-view attention to establish inter-view correspondences and aggregate global scene geometry. This unified design eliminates the need for a separate multi-view fusion module and naturally accommodates a variable number of views, enabling deployment across robotic platforms with diverse camera configurations.

MAGP supports flexible observation configurations. For each view $i$, available camera parameters $(K_i,R_i,t_i)$ and depth $D_i$ are encoded as $z_i^C=\varepsilon_C(K_i,R_i,t_i)$ and $z_i^D=\varepsilon_D(D_i)$. $\varepsilon_C(\cdot)$ is a lightweight MLP and $\varepsilon_D(\cdot)$ is a patch embedding. Missing observations are represented by shared learnable tokens $z_{\varnothing}^{C}$ and $z_{\varnothing}^{D}$. The camera token is concatenated with the image tokens, whereas the depth tokens are added to the image tokens as $\widetilde z_i^I=z_i^I+z_i^D$. Unlike scene-wise normalization, which removes metric scale, MAGP applies a scene-independent normalization factor $\mu$ to camera translation and depth observations. This uniform linear rescaling places metric quantities within a consistent numerical range, thereby improving numerical stability during optimization without introducing scene-dependent scale distortion.


The DPT head decodes metric geometry tokens into metric scale depth maps, while the camera head estimates the corresponding intrinsics and extrinsics. These predictions are jointly back-projected into a shared metric coordinate system to recover scene geometry.

\paragraph{Metric to Action Grounding.}
To inject metric geometry into the embodied policy, we introduce a residual cross-attention module that grounds visual-language features in metric scale geometric representations. Let $\mathbf{H}_{\mathrm{vl}}\in\mathbb{R}^{L\times C}$ denote the visual-language tokens produced by Qwen3-VL-4B-Instruct~\cite{bai2025qwen3}, and $\mathbf{H}_{\mathrm{geo}}\in\mathbb{R}^{M\times C_g}$ denote the metric geometry tokens extracted by MAGP.

The two representations are first projected into a shared latent space:
\begin{equation}
    \mathbf{Q}=\mathbf{H}_{\mathrm{vl}}\mathbf{W}_{Q},
    \mathbf{K}=\mathbf{H}_{\mathrm{geo}}\mathbf{W}_{K},
    \mathbf{V}=\mathbf{H}_{\mathrm{geo}}\mathbf{W}_{V},
\end{equation}
where $\mathbf{W}_{Q}$, $\mathbf{W}_{K}$, and $\mathbf{W}_{V}$ are learnable projection matrices.

Then, the visual-language tokens serve as queries, while the metric geometry tokens provide keys and values:
\begin{equation}
    \mathbf{H}_{\mathrm{fuse}}=\mathbf{H}_{\mathrm{vl}}+CrossAttn\left(\mathbf{Q},\mathbf{K},\mathbf{V}\right).
\end{equation}

This residual fusion enables task-relevant metric cues, including object dimensions, spatial distances, and scene geometry, to be selectively integrated into the visual-language representation. The resulting features preserve the semantic reasoning capability of Qwen3-VL-4B-Instruct while providing scale-consistent geometric grounding for action prediction.

\subsection{Robust Metric-Aware Geometry Optimization}

Providing camera parameters and depth observations alone is insufficient for reliable metric geometry recovery, as the reconstruction model favors visual priors over the physical constraints encoded by these observations. This problem is further complicated by heterogeneous sensing configurations and variations in depth availability and density. MAGP addresses these challenges through the following strategies.


\paragraph{Metric Scale Equivariant Augmentation.}

A metric-aware model should respond proportionally to changes in the scale specified by metric observations. To introduce this attribute while maintaining a stable numerical range, we parameterize the augmentation scale as $\alpha \sim abs(\mathcal{N}(\mu,1))$. For each training scene, camera translation, depth observations, and geometric supervision are jointly rescaled as
\begin{equation}
    \tilde{\mathbf{t}}_i=\alpha\mathbf{t}_i,
    \tilde{\mathbf{D}}_i=\alpha\mathbf{D}_i,
    \tilde{\mathbf{P}}_i=\alpha\mathbf{P}_i,
    \label{eq:metric_scale_aug}
\end{equation}

According to Eq.~\eqref{eq_back_projection}, this transformation preserves image projections while scaling the physical scene geometry by $\alpha$. By associating identical visual observations with different metric scales, the augmentation prevents visual appearance from determining reconstruction scale and encourages MAGP to follow the physical constraints encoded by camera translation and depth.

\paragraph{Flexible Metric Conditioning.}
Robotic platforms exhibit heterogeneous sensing configurations: camera parameters may be available only for selected views, while depth observations can be dense, sparse, or entirely missing. Training with a fixed combination of metric observations can therefore cause over-dependence on a specific sensor configuration and limit cross-platform generalization.

To simulate diverse observation patterns, MAGP independently varies camera and depth availability for each view. Specifically,
\begin{equation}
    m_i^{C},m_i^{D}
    \sim
    \operatorname{Bernoulli}(0.5),
\end{equation}
where $m=1$ retains the corresponding observation and $m=0$ replaces it with a shared learnable placeholder. The camera representation is defined as
\begin{equation}
    \tilde{\mathbf{z}}_i^{C}
    =
    m_i^{C}\mathbf{z}_i^{C}
    +
    \left(1-m_i^{C}\right)\mathbf{z}_{\varnothing}^{C}.
    \label{eq:camera_conditioning}
\end{equation}

To account for varying depth density, a stochastic spatial mask $\mathbf{S}_i$ is further applied:
\begin{equation}
    \tilde{\mathbf{D}}_i^*
    =
    \mathbf{S}_i\odot\tilde{\mathbf{D}}_i,
\end{equation}
where $\odot$ denotes element-wise multiplication. The depth representation is then constructed as
\begin{equation}
    \tilde{\mathbf{z}}_i^{D}
    =
    m_i^{D}\varepsilon_D\!\left(\tilde{\mathbf{D}}_i^*\right)
    +
    \left(1-m_i^{D}\right)\mathbf{z}_{\varnothing}^{D}.
    \label{eq:depth_conditioning}
\end{equation}


By independently varying camera availability, depth availability, and depth density across views, MAGP unifies camera-conditioned, depth-conditioned, jointly conditioned, and partially observed settings within a single model. It leverages available metric observations to recover metric geometry and, when such observations are sparse or missing, complements them with the remaining observations and multi-view geometric constraints. This flexible conditioning mechanism reduces reliance on fixed sensor configurations and improves robustness to heterogeneous sensing conditions.

\paragraph{Training Objective.}
MAGP jointly optimizes relative and metric objectives. As defined in Eqs. (\ref{eq_rel_loss}) and (\ref{eq_metric_loss}), respectively, the relative supervision $\mathcal{L}_{\mathrm{rel}}$ preserves fine-grained geometry and cross-view consistency, whereas the metric supervision $\mathcal{L}_{\mathrm{metric}}$ supervises reconstruction in metric scale. Since metric supervision is valid only when sufficient scale evidence is available, we introduce
\begin{equation}
    m_{\mathrm{metric}}=\mathbb{I}\left[N_{\mathrm{extrinsics}}\geq 2\;\lor\;N_{\mathrm{depth}}\geq 1\right],
    \label{eq_metric_validity}
\end{equation}
where \(N_{\mathrm{extrinsics}}\) and \(N_{\mathrm{depth}}\) denote the numbers of views with valid camera extrinsics and depth observations, respectively. Thus, \(m_{\mathrm{metric}}=1\) when either a metric multi-view baseline or at least one depth observation is available.

Using the \(\ell_1\) discrepancy \(d(\mathbf{x},\mathbf{y})=\lVert\mathbf{x}-\mathbf{y}\rVert_1\), the final objective is
\begin{equation}
    \mathcal{L}_{\mathrm{geo}}
    =
    \mathcal{L}_{\mathrm{rel}}
    +
    m_{\mathrm{metric}}\mathcal{L}_{\mathrm{metric}}.
    \label{eq_joint_geometry_loss}
\end{equation}

\begin{table}
    \centering
    \setlength{\tabcolsep}{0.5mm}
    \small
    \begin{tabular}{lVcccccc}
    \toprule[1.2pt]
        \multirow{2}{*}{\textbf{Method}} & \multicolumn{3}{c}{\textbf{Depth}} & \multicolumn{3}{c}{\textbf{Points}}\\
         & ${\delta _{1.03}}(\uparrow)$ & ${\delta _{1.25}}(\uparrow)$ & \textit{Abs}$(\downarrow)$ & ${\delta _{1.03}}(\uparrow)$ & ${\delta _{1.25}}(\uparrow)$ & \textit{Abs}$(\downarrow)$\\
        \rowcolor{sectiongray}
        \multicolumn{7}{l}{Input: Image} \\
        VGGT & 51.77 & 95.58 & 3.88 & 67.42 & 97.10 & 2.09\\
        MA & 51.92 & 96.99 & 4.97 & 68.53 & 98.13 & 2.17\\
        DA3 & 56.82 & 97.89 & 3.96 & 67.27 & 98.30 & 2.00\\
        VGGT-$\Omega$ & 66.40 & 98.51 & 4.04 & 82.91 & \textbf{99.17} & 2.00\\
        MAGP & \textbf{74.23} & \textbf{99.05} & \textbf{0.75} & \textbf{84.27} & \textbf{99.17} & \textbf{0.40}\\
        \rowcolor{sectiongray}
        \multicolumn{7}{l}{Input: Image, Camera Parameter} \\
        MA & 50.22 & 93.25 & 2.31 & 57.34 & 93.19 & 1.03\\
        DA3 & 62.65 & 98.28 & 3.92 & 72.11 & 98.58 & 2.01\\
        MAGP & \textbf{84.87} & \textbf{99.12} & \textbf{0.12} & \textbf{87.21} & \textbf{99.28} & \textbf{0.07}\\
        \rowcolor{sectiongray}
        \multicolumn{7}{l}{Input: Image, Depth} \\
        MA & 40.00 & 93.30 & 0.29 & 49.33 & 94.80 & 0.24\\
        MAGP & \textbf{73.70} & \textbf{99.15} & \textbf{0.25} & \textbf{83.23} & \textbf{99.28} & \textbf{0.13}\\
        \rowcolor{sectiongray}
        \multicolumn{7}{l}{Input: Image, Camera Parameter, Depth} \\
        MA & 57.50 & 93.96 & 0.21 & 58.11 & 95.36 & 0.12\\
        MAGP & \textbf{84.81} & \textbf{99.16} & \textbf{0.11} & \textbf{87.34} & \textbf{99.34} & \textbf{0.06}\\
    \bottomrule[1.2pt]
    \end{tabular}
    \caption{Comparison of reconstruction accuracy under different settings. Results are averaged over three benchmarks.}
    \label{tab_compare_magp}
\end{table}

\section{Experiments}

We evaluate MAGP on both reconstruction and robotic benchmarks, assessing its metric reconstruction capability and its effectiveness for robotics.

\subsection{Evaluation on 3D Reconstruction Task}

\paragraph{Implementation Details.}

MAGP is initialized from pretrained DA3-Giant model and trained on a mixture of multi-view reconstruction datasets using a two-stage view curriculum. Details are provided in the appendix.

\paragraph{Benchmark and Metric.}
We evaluate metric geometry reconstruction on ETH3D~\cite{schops2017multi}, MegaDepth~\cite{li2018megadepth}, and ScanNet++~\cite{yeshwanth2023scannet++}. Performance is measured using $\delta_{1.03}$, $\delta_{1.25}$, and absolute error (\textit{Abs}). The $\delta$ metrics report the inlier ratios under relative thresholds of $1.03$ and $1.25$, while \textit{Abs} measures the mean absolute error in meters. 


\paragraph{Comparison with State-of-the-Art Methods.}

\begin{table}[!t]
    \centering
    \setlength{\tabcolsep}{1mm}
    \small
    \begin{tabular}{cVccccc}
    \toprule[1.2pt]
        \multirow{2}{*}{\textbf{Input}} & \multirow{2}{*}{\textbf{MSEA}} & \multicolumn{2}{c}{\textbf{Depth}} & \multicolumn{2}{c}{\textbf{Points}} \\
         & & ${\delta _{1.03}}(\uparrow)$ & \textit{Abs}$(\downarrow)$ & ${\delta _{1.03}}(\uparrow)$ & \textit{Abs}$(\downarrow)$ \\
        \midrule
        \multirow{2}{*}{$[\mathcal{R}|0.9t]$} & $\times$ & 69.58 & 0.11 & 74.57 & 0.09\\
         & \cellcolor{sectiongray}$\checkmark$ & \cellcolor{sectiongray}\textbf{78.68} & \cellcolor{sectiongray}\textbf{0.05} & \cellcolor{sectiongray}\textbf{83.53} & \cellcolor{sectiongray}\textbf{0.04}\\
        \multirow{2}{*}{$[\mathcal{R}|t]$} & $\times$ & 70.72 & 0.06 & 75.97 & 0.06\\
         & \cellcolor{sectiongray}$\checkmark$ & \cellcolor{sectiongray}\textbf{78.86} & \cellcolor{sectiongray}\textbf{0.04} & \cellcolor{sectiongray}\textbf{83.90} & \cellcolor{sectiongray}\textbf{0.04}\\
        \multirow{2}{*}{$[\mathcal{R}|1.1t]$} & $\times$ & 71.60 & 0.07 & 77.08 & 0.07\\
         & \cellcolor{sectiongray}$\checkmark$ & \cellcolor{sectiongray}\textbf{79.21} & \cellcolor{sectiongray}\textbf{0.04} & \cellcolor{sectiongray}\textbf{83.94} & \cellcolor{sectiongray}\textbf{0.04}\\
        \multirow{2}{*}{$[\mathcal{R}|2t]$} & $\times$ & 53.74 & 0.45 & 53.21 & 0.36\\
         & \cellcolor{sectiongray}$\checkmark$ & \cellcolor{sectiongray}\textbf{73.84} & \cellcolor{sectiongray}\textbf{0.06} & \cellcolor{sectiongray}\textbf{80.21} & \cellcolor{sectiongray}\textbf{0.05}\\
        \midrule
        \multirow{2}{*}{0.9$\mathcal{D}$} & $\times$ & 70.82 & 0.17 & 76.43 & 0.14\\
         & \cellcolor{sectiongray}$\checkmark$ & \cellcolor{sectiongray}\textbf{75.84} & \cellcolor{sectiongray}\textbf{0.06} & \cellcolor{sectiongray}\textbf{83.02} & \cellcolor{sectiongray}\textbf{0.06}\\
        \multirow{2}{*}{$\mathcal{D}$} & $\times$ & 70.73 & 0.07 & 75.88 & 0.07\\
         & \cellcolor{sectiongray}$\checkmark$ & \cellcolor{sectiongray}\textbf{76.43} & \cellcolor{sectiongray}\textbf{0.06} & \cellcolor{sectiongray}\textbf{83.26} & \cellcolor{sectiongray}\textbf{0.05}\\
        \multirow{2}{*}{1.1$\mathcal{D}$} & $\times$ & 70.30 & 0.13 & 75.82 & 0.12\\
         & \cellcolor{sectiongray}$\checkmark$ & \cellcolor{sectiongray}\textbf{76.07} & \cellcolor{sectiongray}\textbf{0.07} & \cellcolor{sectiongray}\textbf{83.56} & \cellcolor{sectiongray}\textbf{0.06}\\
        \multirow{2}{*}{2$\mathcal{D}$} & $\times$ & 69.83 & 0.63 & 74.92 & 0.53\\
         & \cellcolor{sectiongray}$\checkmark$ & \cellcolor{sectiongray}\textbf{76.18} & \cellcolor{sectiongray}\textbf{0.09} & \cellcolor{sectiongray}\textbf{83.31} & \cellcolor{sectiongray}\textbf{0.08}\\
    \bottomrule[1.2pt]
    \end{tabular}
    \caption{Ablation of Metric Scale Equivariant Augmentation (MSEA). Translation $t$ and depth $\mathcal{D}$ are independently rescaled while visual observations remain fixed.}
    \label{tab_compare_msea}
\end{table}

As shown in Table~\ref{tab_compare_magp}, MAGP consistently outperforms VGGT, MapAnything (MA), Depth Anything 3 (DA3), and VGGT-$\Omega$ across all observation settings. With images alone, it achieves $\delta_{1.03}$ scores of 74.23\% for depth and 84.27\% for points, surpassing DA3 by 17.41\% and 17.00\%, respectively, demonstrating strong relative geometry reconstruction.

When camera parameters or depth observations are available, MAGP substantially improves metric accuracy while preserving relative geometry. Using camera parameters, it achieves \textit{Abs} errors of $0.12\mathrm{m}$ for depth and $0.07\mathrm{m}$ for points; using depth observations, the errors are $0.25\mathrm{m}$ and $0.13\mathrm{m}$. With both observations, MAGP attains the best performance, reducing the errors to $0.11\mathrm{m}$ and $0.06\mathrm{m}$.

\begin{figure*}[t]
\centering
\includegraphics[width=0.95\linewidth]{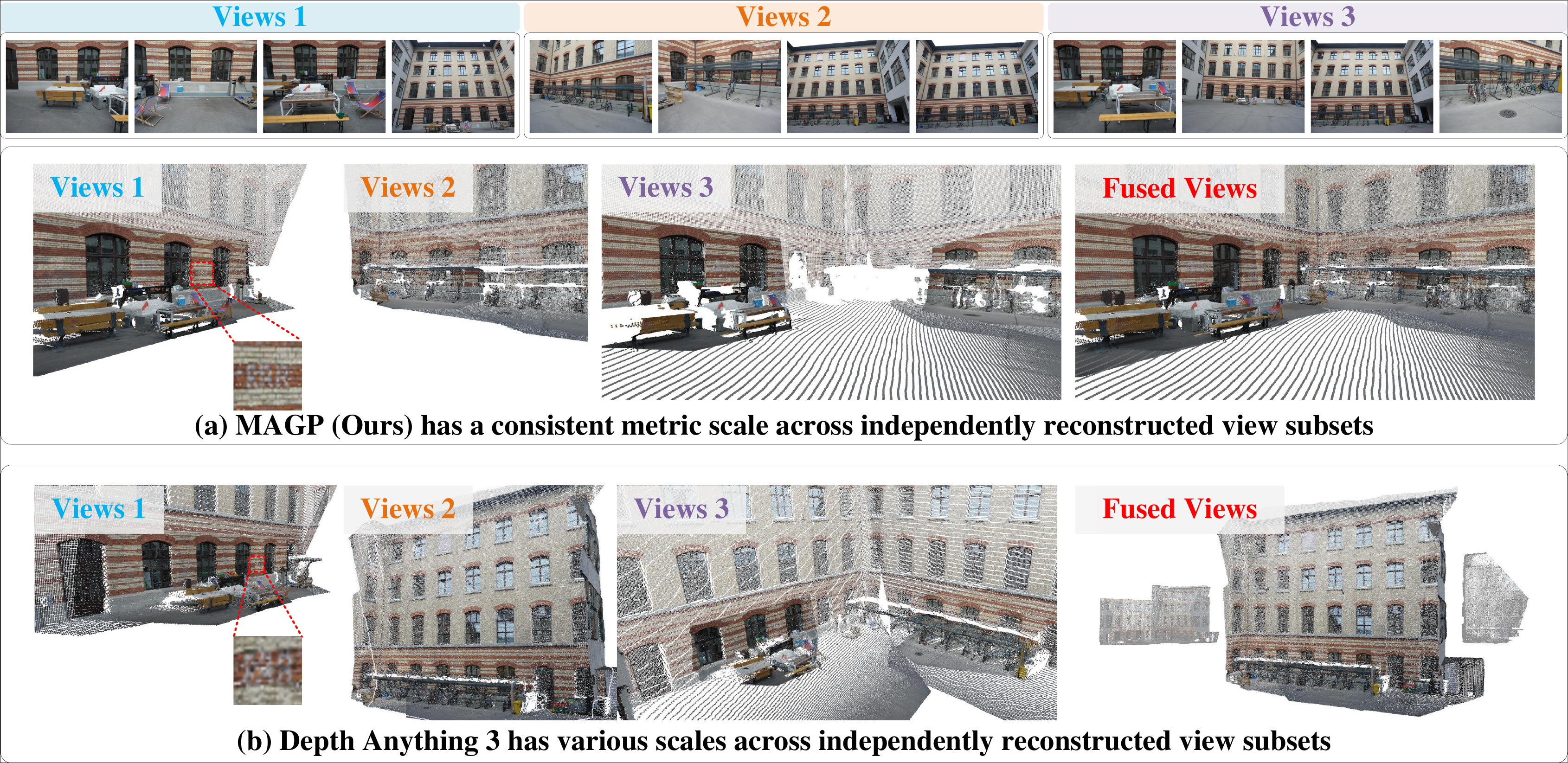}
\caption{Scale consistency across three independently reconstructed subsets. MAGP achieves consistency reconstruction.}
\label{fig_pr_vis}
\end{figure*}

Fig.~\ref{fig_pr_vis} further demonstrates cross-batch scale consistency. MAGP maintains a consistent metric scale across independently reconstructed view subsets, enabling coherent fusion, whereas DA3 produces subset-dependent scales that lead to severe misalignment and fragmented geometry.



\paragraph{Ablation Studies.}

Table~\ref{tab_compare_msea} evaluates the effectiveness of MSEA on ScanNet++. MSEA enables the model to consistently follow scale variations, whereas removing it causes substantial degradation and weak responsiveness to the provided metric observations. Together with Fig.~\ref{fig_scale_equivariance}, these results show that MSEA is essential for preventing reliance on visual priors and achieving metric scale equivariant reconstruction.

\begin{table}[!t]
    \centering
    \small
    \begin{tabular}{cVcccc}
    \toprule[1.2pt]
        \multirow{2}{*}{$\mathcal{L}_{geo}$} & \multicolumn{2}{c}{\textbf{Depth}} & \multicolumn{2}{c}{\textbf{Points}} \\
         & ${\delta _{1.03}}(\uparrow)$ & \textit{Abs}$(\downarrow)$ & ${\delta _{1.03}}(\uparrow)$ & \textit{Abs}$(\downarrow)$ \\
        \midrule
        $\mathcal{L}_{\mathrm{rel}}$ & 63.53 & 2.54 & 73.90 & 1.67 \\
        $\mathcal{L}_{\mathrm{metric}}$ & 77.44 & 0.16 & 81.97 & 0.10 \\
        $\mathcal{L}_{\mathrm{rel}}$\&$\mathcal{L}_{\mathrm{metric}}$ & \textbf{80.67} & \textbf{0.15} & \textbf{83.30} & \textbf{0.09} \\
    \bottomrule[1.2pt]
    \end{tabular}
    \caption{Ablation of training objective $\mathcal{L}_{geo}$. Joint relative and metric supervision improves geometric accuracy while maintaining metric scale consistency.}
    \label{tab_ab_loss}
\end{table}


Table~\ref{tab_ab_loss} isolates the effects of different training objectives on ScanNet++. The relative objective $\mathcal{L}_{\mathrm{rel}}$ preserves accurate geometric structure but does not explicitly constrain the reconstruction to a metric scale. In contrast, $\mathcal{L}_{\mathrm{metric}}$ enables metric geometry recovery, although its relative accuracy remains below that achieved by the joint objective. Jointly optimizing $\mathcal{L}_{\mathrm{rel}}$ and $\mathcal{L}_{\mathrm{metric}}$ yields the best overall performance, confirming their complementarity: relative supervision preserves fine-grained geometric structure, whereas metric supervision anchors the reconstruction to a consistent metric scale.


\begin{table}[!t]
    \centering
    \small
    \begin{tabular}{cVcccc}
    \toprule[1.2pt]
        \multirow{2}{*}{$\mu$} & \multicolumn{2}{c}{\textbf{Depth}} & \multicolumn{2}{c}{\textbf{Points}} \\
         & ${\delta _{1.03}}(\uparrow)$ & \textit{Abs}$(\downarrow)$ & ${\delta _{1.03}}(\uparrow)$ & \textit{Abs}$(\downarrow)$ \\
        \midrule
        5 & 80.74 & 0.15 & \textbf{85.53} & \textbf{0.08} \\
        10 & \textbf{82.99} & \textbf{0.13} & 84.47 & \textbf{0.08} \\
        15 & 80.90 & 0.16 & 81.24 & 0.09 \\
    \bottomrule[1.2pt]
    \end{tabular}
    \caption{Ablation of the normalization factor $\mu$. $\mu$ serves both as the normalization factor for metric inputs and as the mean of the Gaussian distribution used in MSEA.}
    \label{tab_ab_mean}
\end{table}

\begin{table*}[!t]
    \centering
    \small
    \begin{tabular}{lVcccccccVc}
    \toprule[1.2pt]
        \textbf{Method} & \textbf{Camera} & \textbf{Robot} & \textbf{Language} & \textbf{Light} & \textbf{Background} & \textbf{Noise} & \textbf{Layout} & \textbf{Total} \\
        \midrule
        OpenVLA & 0.8 & 3.5 & 23.0 & 8.1 & 34.8 & 15.2 & 28.5 & 15.6 \\
        OpenVLA-OFT & 56.4 & 31.9 & 79.5 & 88.7 & 93.3 & 	75.8 & 74.2 & 69.6 \\
        $\pi_0$ & 13.8 & 6.0 & 58.8 & 85.0 & 81.4 & 79.0 & 68.9 & 53.6 \\
        $\pi_0$-Fast & 65.1 & 21.6 & 61.0 & 73.2 & 73.2 & 74.4 & 68.8 & 61.6 \\
        $\pi_{0.5}$ & 64.3 & 57.2 & 82.8 & 94.2 & 94.0 & 79.6 & 78.2 & 77.0 \\
        RIPT-VLA & 55.2 & 31.2 & 77.6 & 88.4 & 91.6 & 73.5 & 74.2 & 68.4 \\
        \midrule
        GR00T & 60.9 & 49.0 & 87.6 & 92.5 & \textbf{95.5} & 81.3 & 78.5 & 76.4 \\
        GR00T$^1$ & \textbf{65.9} & 56.8 & \textbf{88.1} & 94.1 & 94.5 & 83.6 & 79.0 & 79.0 \\
        GR00T$^2$ & 62.7 & \textbf{63.9} & 88.0 & \textbf{96.4} & 94.7 & \textbf{86.1} & \textbf{79.3} & \textbf{80.3$_{+3.9}$} \\
    \bottomrule[1.2pt]
    \end{tabular}
    \caption{Zero-shot performance comparison on LIBERO-Plus. Task success rates (\%) are reported. $^{1}$ denotes the use of relative geometry, while $^{2}$ denotes the use of metric geometry (MAGP).}
    \label{tab_compare_liberoplus}
\end{table*}

We set the normalization factor to $\mu=10$ based on the characteristic spatial scales of environments relevant to robotic perception. Hypersim captures complex indoor scenes, whereas CO3Dv2 provides object-centric observations of everyday objects from diverse viewpoints, jointly covering scene-level understanding and local interaction. Their mean scene depths are $9.39\mathrm{m}$ and $10.92\mathrm{m}$, respectively, supporting a representative scale of approximately $10\mathrm{m}$. Table~\ref{tab_ab_mean} further shows that this physically grounded choice provides a stable trade-off across depth and point reconstruction metrics.


\begin{table}
    \centering
    \small
    \setlength{\tabcolsep}{1mm}
    \begin{tabular}{lVccccVc}
    \toprule[1.2pt]
        \textbf{Method} & \textbf{Spatial} & \textbf{Object} & \textbf{Goal} & \textbf{Long} & \textbf{Avg.} \\
        \midrule
        $\pi_0$ & 96.8 & 98.8 & 95.8 & 85.2 & 94.1 \\
        $\pi_{0.5}$ & 98.8 & 98.2 & 98.0 & 92.4 & 96.9 \\
        OpenVLA & 84.7 & 88.4 & 79.2 & 53.7 & 76.5 \\
        OpenVLA-OFT & 97.6 & 98.4 & 97.9 & 94.5 & 97.1 \\
        Motus & 96.8 & 99.8 & 96.6 & \textbf{97.6} & 97.7 \\
        DepthVLA$^1$ & 96.4 & 98.0 & 95.8 & 89.2 & 94.9 \\
        Fast-WAM & 98.2 & \textbf{100.0} & 97.0 & 95.2 & 97.6 \\
        \midrule
        StarVLA-$\pi$ & 98.2 & 99.8 & 98.2 & 90.8 & 96.8 \\
        StarVLA-$\pi$$^1$ & 98.4 & 99.4 & 98.6 & 92.8 & 97.3$_{+0.5}$ \\
        StarVLA-$\pi$$^2$ & 99.4 & 98.4 & \textbf{98.8} & 95.6 & 98.1$_{+1.3}$ \\
        \midrule
        GR00T & 99.2 & 98.8 & 98.0 & 94.0 & 97.5 \\
        GR00T$^1$ & 99.4 & 98.4 & \textbf{98.8} & 95.6 & 98.1$_{+0.6}$ \\
        GR00T$^2$ & \textbf{99.8} & 99.2 & 98.4 & 96.6 & \textbf{98.5$_{+1.0}$} \\
    \bottomrule[1.2pt]
    \end{tabular}
    \caption{Performance comparison on LIBERO. Task success rates (\%) are reported. $^{1}$ denotes the use of relative geometry, while $^{2}$ denotes the use of metric geometry (MAGP).}
    \label{tab_compare_libero}
\end{table}


\subsection{Evaluation on Robotic Manipulation Task}

\paragraph{Implementation Details.}

Our policy is trained from scratch for 100K iterations on LIBERO, while the RoboTwin policy is initialized from the pretrained DM0.5~\cite{dm05} weights and trained for 150K iterations. Details are provided in the appendix.

\begin{table}
    \centering
    \small
    \begin{tabular}{lVccVc}
    \toprule[1.2pt]
        \textbf{Method} & \textbf{Clean} & \textbf{Rand.} & \textbf{Avg.}\\
        \midrule
        $\pi_{0}$ & 65.92 & 58.40 & 62.16 \\
        $\pi_{0.5}$ & 82.74 & 76.76 & 79.75 \\
        X-VLA & 72.80 & 72.84 & 72.82 \\
        Motus & 88.66 & 87.02 & 87.84 \\
        ABot-M0$^1$ & 86.06 & 85.08 & 85.57 \\
        \midrule
        GR00T & 82.06 & 81.32 & 81.69 \\
        GR00T$^1$ & 86.70 & 83.64 & 85.17$_{+3.48}$ \\
        GR00T$^2$ & 87.46 & 88.44 & 87.95$_{+6.26}$ \\
        \midrule
        DM0.5 & 92.72 & 91.52 & 92.12 \\
        DM0.5$^2$ & \textbf{94.06} & \textbf{93.76} & \textbf{93.91$_{+1.79}$} \\
    \bottomrule[1.2pt]
    \end{tabular}
    \caption{Performance comparison on RoboTwin. Task success rates (\%) are reported. $^{1}$ denotes the use of relative geometry, while $^{2}$ denotes the use of metric geometry (MAGP).}
    \label{tab_compare_robotwin}
\end{table}

\paragraph{Benchmark and Evaluation Protocol.}


We evaluate robotic manipulation on LIBERO~\cite{liu2023libero} and RoboTwin~\cite{chen2025robotwin}, and assess zero-shot robustness on LIBERO-Plus~\cite{fei2026libero} without additional fine-tuning. We compare MAGP with representative state-of-the-art methods~\cite{black2026pi0visionlanguageactionflowmodel, kim2024openvla, kim2025finetuningvisionlanguageactionmodelsoptimizing, bi2025motusunifiedlatentaction, yuan2025depthvla, yuan2026fastwamworldactionmodels, black2025pi_, bjorck2025gr00t,pertsch2025fastefficientactiontokenization, ye2026starvla}. Details are provided in the appendix.




\paragraph{Performance comparison on LIBERO and RoboTwin.}

As shown in Tables~\ref{tab_compare_libero} and~\ref{tab_compare_robotwin}, metric geometry consistently improves manipulation performance. On LIBERO, the average success rate increases from 96.8\% to 98.1\% for StarVLA-$\pi$ and from 97.5\% to 98.5\% for GR00T. On RoboTwin, it improves GR00T from 81.69\% to 87.95\% and DM0.5 from 92.12\% to 93.91\%, demonstrating the effectiveness of metric-consistent geometry across models and benchmarks.

\paragraph{Zero-shot Performance on LIBERO-PLUS.}

Table~\ref{tab_compare_liberoplus} reports zero-shot results on LIBERO-Plus. GR00T$^{2}$ achieves the best overall success rate of 80.3\%, outperforming GR00T and GR00T$^{1}$ by 3.9\% and 1.3\%, respectively. The largest gain occurs under robot-state perturbations, from 49.0\% to 63.9\%, with further improvements under light, noise, and layout shifts. These results show that metric geometry improves robustness to embodiment and spatial shifts.



\section{Conclusion}


We propose MAGP, an end-to-end and plug-and-play framework that reconstructs metric geometry for downstream robotic policies. Extensive experiments on 3D reconstruction and robotic manipulation benchmarks demonstrate its effectiveness and consistent advantages.

\bibliography{aaai2027}


\clearpage
\appendix
\section*{Appendix}

\section{3D Reconstruction Task}


\subsection{Training Datasets}

MAGP is trained on a large and diverse collection of real-world and synthetic datasets, including UnrealStereo4K, WildRGB-D, Spring, ParallelDomain4D, MegaDepth, MPSD, Dynamic Replica, CO3Dv2, SAIL-VOS 3D, Hypersim, BlendedMVS, and ScanNet++. These datasets collectively cover indoor and outdoor environments, static and dynamic scenes, and monocular, stereo, and multi-view configurations. Their geometric annotations are obtained from depth sensors, simulation engines, or structure-from-motion and multi-view stereo pipelines, providing diverse supervision for relative- and metric-scale reconstruction.

\subsection{Implementation Details}

MAGP is initialized from the pretrained DA3-Giant model. Training is conducted on 32 NVIDIA H100 GPUs using the AdamW optimizer with a learning rate of $5\times10^{-6}$ and a weight decay of $0.05$. The learning rate follows a cosine decay schedule throughout optimization. Mixed-precision training and gradient checkpointing are employed to improve computational efficiency and reduce memory consumption. A two-stage view curriculum is adopted to progressively strengthen multi-view geometric reasoning. In the first stage, MAGP is trained for 240K optimization steps, with the number of views per scene varying from 2 to 4. The resulting checkpoint is then further optimized for 38K steps, with the number of views varying from 2 to 18.

\subsection{Model Complexity and Inference Efficiency}

Table~\ref{tab_ab_latency} compares the model complexity and inference efficiency of MAGP with the DA3 baseline. Both methods employ the same backbone and prediction head. MAGP introduces only $0.001$B additional parameters, which originate from a lightweight depth encoder implemented as a patch-embedding layer to project the input depth map into patch-level representations. Inference latency is measured on a single NVIDIA A800 GPU under identical experimental settings for both methods. Compared with DA3, MAGP increases the latency only marginally, from $152\mathrm{ms}$ to $154\mathrm{ms}$. These results demonstrate that MAGP enables metric-aware geometry perception with negligible parameter and computational overhead, facilitating its efficient and plug-and-play integration into robotic manipulation policies.

\begin{figure*}[t]
\centering
\includegraphics[width=0.99\linewidth]{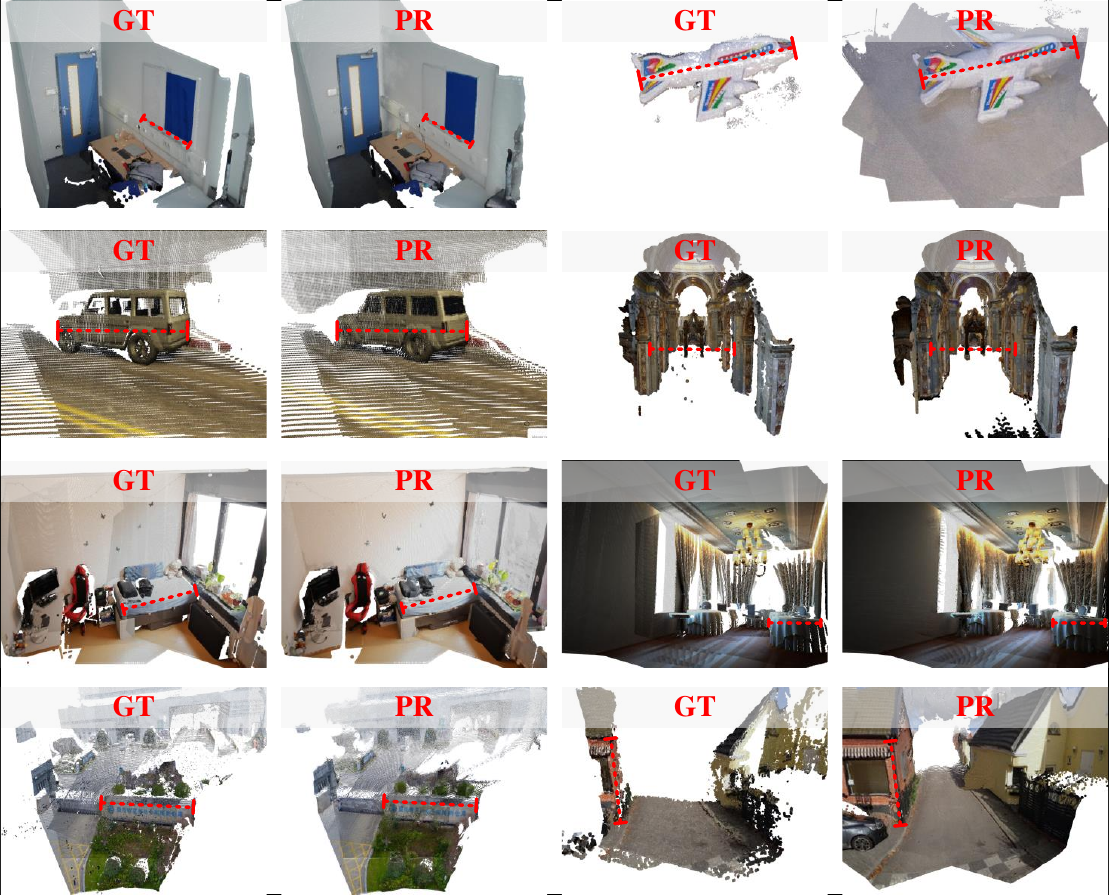}
\caption{Qualitative presentation of metric geometry reconstruction. Across diverse indoor and outdoor scenes, MAGP accurately recovers metric geometry whose scene dimensions and spatial extent match the ground truth, while preserving fine-grained geometric structures. In each GT-PR pair, the red dashed lines are drawn with identical lengths at corresponding structures, providing an intuitive visual reference for evaluating the metric consistency between the prediction and the ground truth.}
\label{fig_pr_view}
\end{figure*}

\subsection{Qualitative Results}

Fig.~\ref{fig_pr_view} qualitatively compares the metric geometry reconstructed by MAGP with the ground truth across diverse indoor and outdoor scenes. The predictions closely match the ground truth in overall scene extent, object dimensions, and spatial distances, while preserving fine-grained structures and complex geometric layouts. Notably, this agreement is achieved without post-hoc scale alignment, indicating that MAGP recovers not only accurate relative geometry but also the absolute size of the scene. These results visually confirm that MAGP produces geometrically detailed reconstructions with a consistent metric scale across diverse environments.

\section{Robotic Manipulation Task}

\subsection{Evaluation Benchmarks and Protocols}

\begin{table}
    \centering
    \small
    \begin{tabular}{lVcccVc}
    \toprule[1.2pt]
        \textbf{Method} & \textbf{Backbone} & \textbf{Encoder} & \textbf{Head} & \textbf{Latency} \\
        \midrule
        DA3 & 1.136B & 0.115B & 0.069B & 152ms\\
        MAGP & 1.136B & 0.116B & 0.069B & 154ms\\
    \bottomrule[1.2pt]
    \end{tabular}
    \caption{Model complexity and inference efficiency analysis.}
    \label{tab_ab_latency}
\end{table}

We evaluate robotic manipulation on LIBERO, LIBERO-Plus, and RoboTwin.

For LIBERO, the policy is trained exclusively on demonstrations from the four standard suites: LIBERO-Spatial, LIBERO-Object, LIBERO-Goal, and LIBERO-Long. Each suite contains 10 tasks, and each task is evaluated over 50 independent rollouts, resulting in 2,000 evaluation episodes in total. Performance is reported as the average task success rate.

We further evaluate on RoboTwin, a large-scale benchmark comprising 50 diverse dual-arm manipulation tasks. These tasks cover a broad range of bimanual skills, including object transfer, tool use, stacking, articulated-object manipulation, and coordinated dual-arm operation. Following the standard protocol, each policy is evaluated under both clean and randomized settings. Each task is tested over 100 independent rollouts in each setting, resulting in 10,000 evaluation episodes in total. The randomized setting introduces variations in scene clutter, lighting, backgrounds, tabletop height, and language instructions. We report the average success rate across all tasks for both settings to evaluate manipulation capability and robustness under diverse environmental conditions.

To assess robustness beyond the training distribution, the LIBERO-trained policy is evaluated zero-shot on LIBERO-Plus without additional fine-tuning. LIBERO-Plus expands the original tasks into 10,030 variants through controlled perturbations across seven dimensions: object layouts, camera viewpoints, robot initial states, language instructions, lighting conditions, background textures, and sensor noise. Each variant is evaluated once due to the scale of the benchmark. We report success rates for each perturbation dimension and across the complete benchmark, measuring generalization to changes in geometry, appearance, embodiment, and language.

\subsection{Implementation Details}

For LIBERO, policies are trained for 100K iterations on 16 NVIDIA H100 GPUs with a per-GPU batch size of 4, yielding a global batch size of 64. For RoboTwin, policies are trained for 150K iterations on 32 NVIDIA H100 GPUs with a per-GPU batch size of 1, resulting in a global batch size of 32. The StarVLA-$\pi$ and GR00T policies are trained from scratch, whereas DM0.5 is initialized from its pretrained weights and subsequently fine-tuned on RoboTwin. Since the original training data do not provide camera intrinsics, extrinsics, or depth measurements as metric observations, we use MoGe to estimate depth from the head-camera images and treat the resulting predictions as depth observations for MAGP. Optimization is performed using AdamW with $\beta_1=0.9$, $\beta_2=0.95$, $\epsilon=1\times10^{-8}$, and a weight decay of $1\times10^{-8}$. The learning rate is set to $1\times10^{-5}$ and follows a cosine decay schedule throughout training. No training or fine-tuning is performed on LIBERO-Plus. The LIBERO-trained policy is directly evaluated under the zero-shot setting.

\subsection{Evaluation}

\begin{figure*}[t]
\centering
\includegraphics[width=0.99\linewidth]{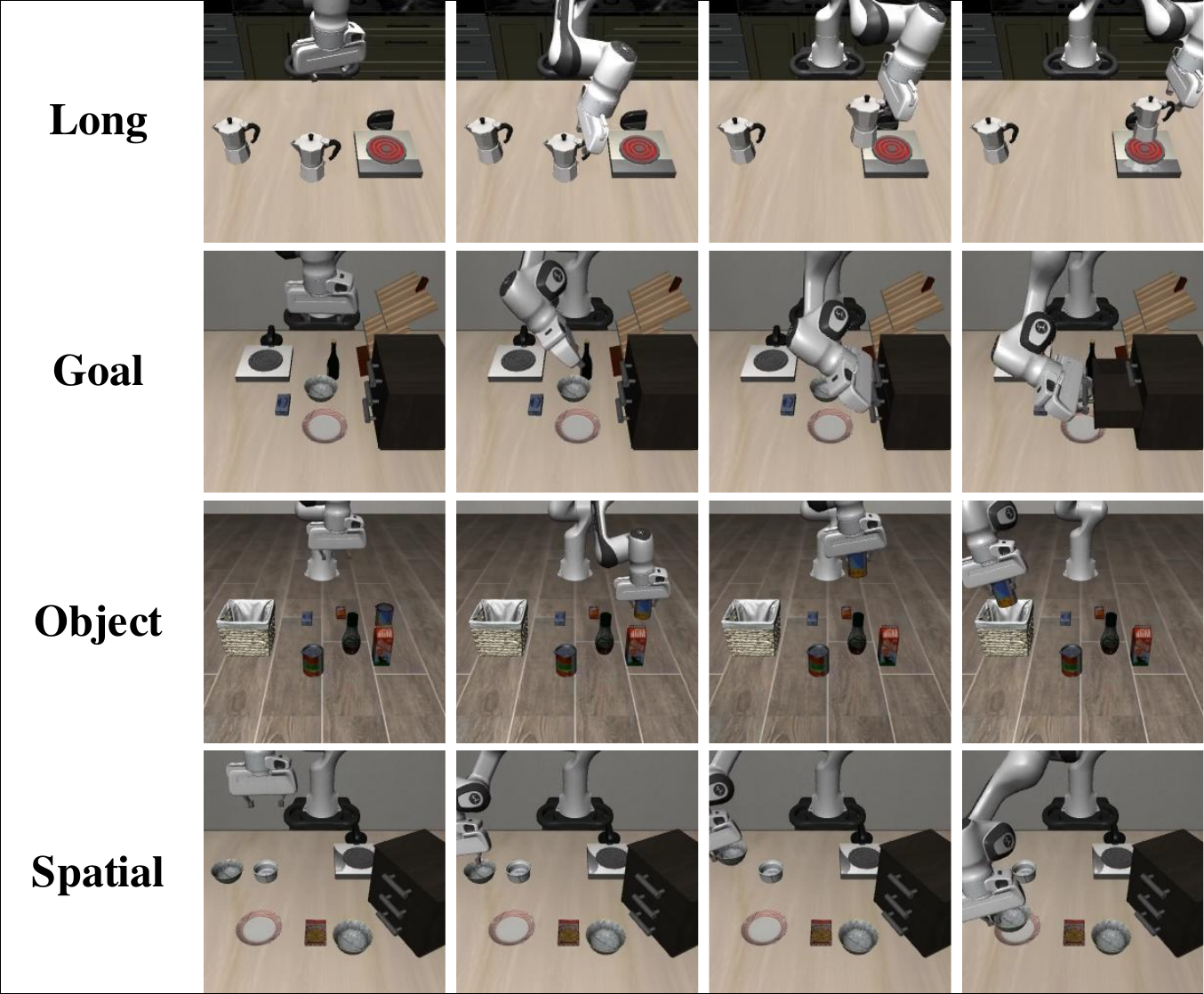}
\caption{Representative Policy Rollouts on LIBERO. LIBERO comprises LIBERO-Long, LIBERO-Goal, LIBERO-Object, and LIBERO-Spatial, which respectively evaluate spatial-relation reasoning, object-centric generalization, goal-conditioned manipulation, and long-horizon task execution. From top to bottom, each row presents a representative rollout from LIBERO-Long, LIBERO-Goal, LIBERO-Object, and LIBERO-Spatial.
}
\label{fig_pr_libero}
\end{figure*}

\paragraph{LIBERO.}
Fig.~\ref{fig_pr_libero} presents representative rollouts from the four LIBERO suites. LIBERO-Spatial evaluates reasoning about relative object positions, LIBERO-Object tests object-centric generalization, LIBERO-Goal focuses on goal-conditioned manipulation, and LIBERO-Long requires long-horizon action execution. Across these settings, the policies accurately localize target objects, approach them with appropriate motion amplitudes, and complete multi-stage interactions. These qualitative results indicate that MAGP provides consistent object dimensions and spatial distances that benefit both fine-grained manipulation and long-horizon task execution.

\begin{figure*}[t]
\centering
\includegraphics[width=0.99\linewidth]{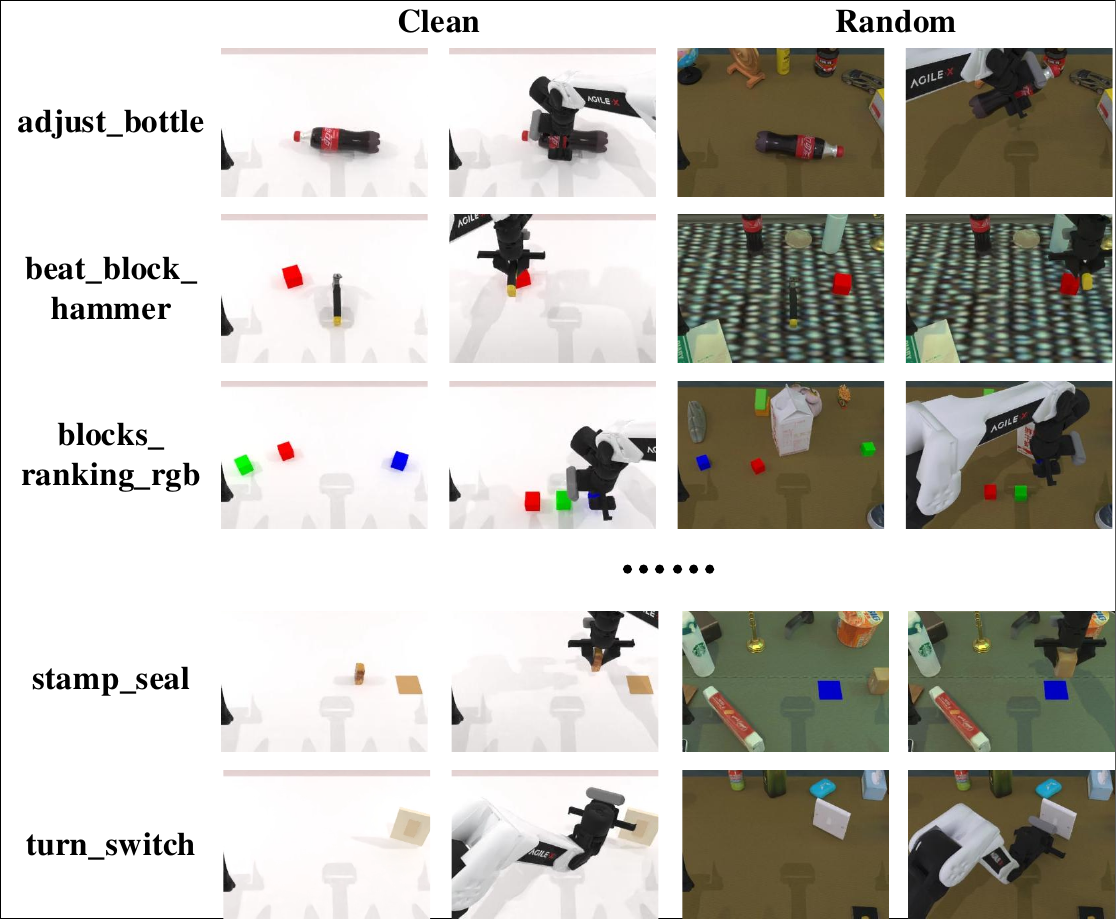}
\caption{Representative policy rollouts under clean and randomized RoboTwin Settings. Each row shows the initial and final states of a representative task. The randomized setting introduces variations in object placement, background, lighting, and scene configuration to evaluate policy robustness under environmental distribution shifts.}
\label{fig_pr_robotwin}
\end{figure*}

\paragraph{RoboTwin.}
As illustrated in Fig.~\ref{fig_pr_robotwin}, the evaluated tasks involve diverse interactions such as object adjustment, tool use, block manipulation, stamping, and switch operation. MAGP enables successful execution not only in clean environments but also under randomized object placements, backgrounds, lighting conditions, and scene configurations. The improvements under these more challenging conditions suggest that metric geometry provides stable spatial grounding when visual appearance and scene layouts vary.

\begin{figure*}[t]
\centering
\includegraphics[width=0.99\linewidth]{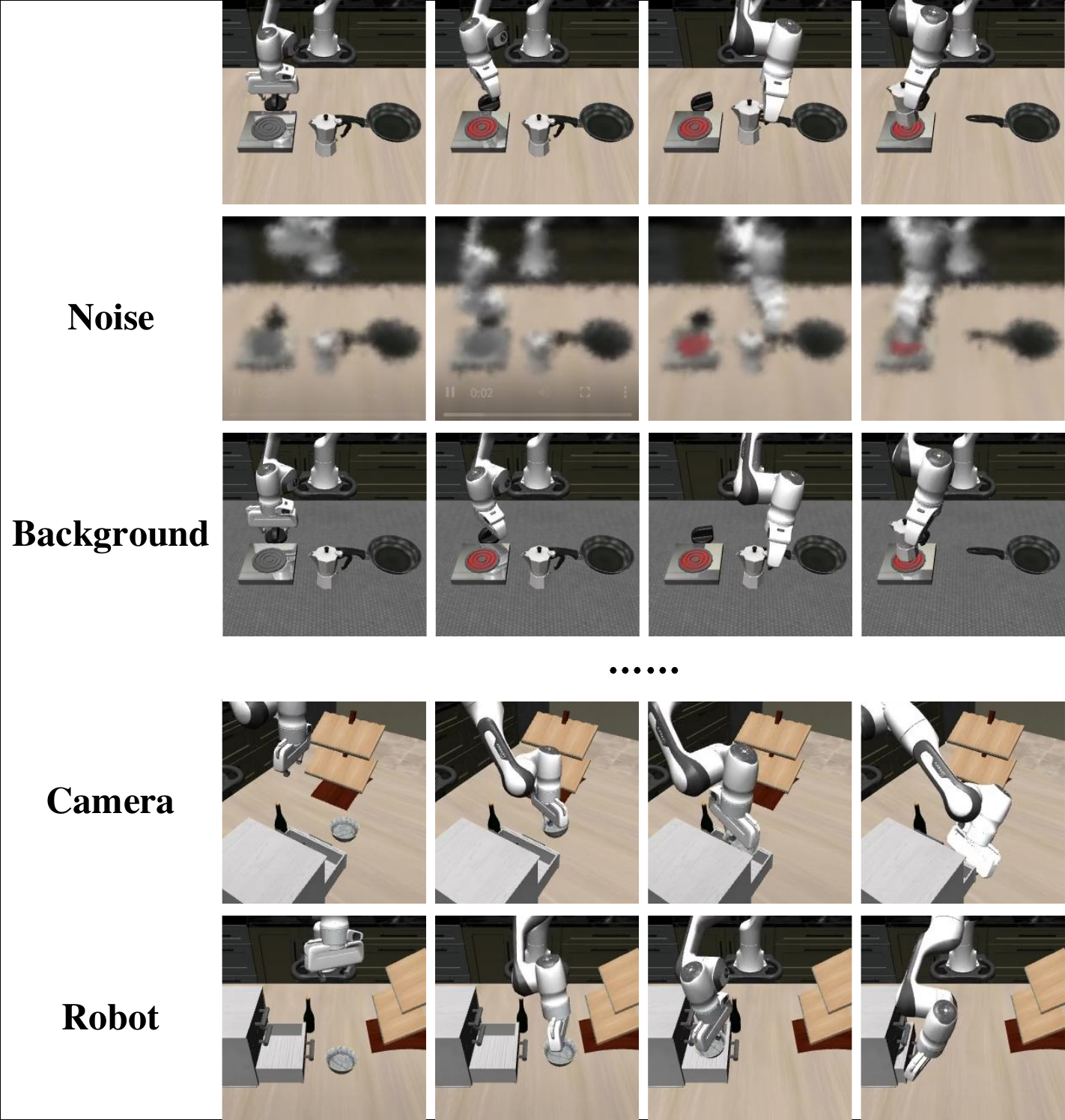}
\caption{Representative zero-shot policy rollouts on LIBERO-Plus. Each row presents a manipulation trajectory under a different distribution shift, demonstrating policy robustness to variations in noise, background, camera and robot states.}
\label{fig_pr_libero_plus}
\end{figure*}

\paragraph{LIBERO-Plus.}
We further evaluate the LIBERO-trained policy on LIBERO-Plus without additional fine-tuning. Fig.~\ref{fig_pr_libero_plus} shows that the policy can retain its manipulation capability under substantial changes in visual appearance, object layout, and robot initialization. These results demonstrate that the geometric representation learned by MAGP remains reliable beyond the training distribution, particularly under shifts that alter embodiment states and spatial relationships.

Overall, the consistent improvements across LIBERO, RoboTwin, and zero-shot LIBERO-Plus demonstrate that MAGP is broadly compatible with different robotic policies and manipulation settings. By providing object dimensions and spatial distances on a consistent metric scale, MAGP strengthens the alignment between geometric perception and action prediction, leading to more accurate and robust manipulation.

\end{document}